\documentclass{article}

\usepackage[nonatbib,preprint]{neurips_2023}

\usepackage{algorithm}
\usepackage{algorithmicx, algpseudocode}
\usepackage[utf8]{inputenc} 
\usepackage[T1]{fontenc}    
\usepackage[bookmarks=false]{hyperref}       
\usepackage{url}            
\usepackage{booktabs}       
\usepackage{amsfonts}       
\usepackage{nicefrac}       
\usepackage{microtype}      
\usepackage{xcolor}         
\usepackage{caption}
\usepackage{graphicx}
\usepackage{bm,multirow}
\usepackage{multicol}
\usepackage{adjustbox}
\usepackage{bm}
\usepackage{wrapfig}

\usepackage{amsmath,amsfonts,bm}

\def\eqref#1{equation~\ref{#1}}

\def\1{\bm{1}}

\DeclareMathAlphabet{\mathsfit}{\encodingdefault}{\sfdefault}{m}{sl}
\SetMathAlphabet{\mathsfit}{bold}{\encodingdefault}{\sfdefault}{bx}{n}

\newcommand{\x}{{\boldsymbol x}}
\renewcommand{\d}{{\boldsymbol d}}

\renewcommand{\v}{{\boldsymbol v}}
\newcommand{\e}{{\boldsymbol e}}
\newcommand{\epsilonb}{{\boldsymbol \epsilon}}

\newcommand{\z}{{\boldsymbol z}}

\newcommand{\Ib}{{\boldsymbol I}}

\newcommand{\Nc}{{\mathcal N}}

\renewcommand{\eqref}[1]{Eq. (\ref{#1})}

\title{Improving Diffusion-based Image Translation using Asymmetric Gradient Guidance}

\author{%
  Gihyun Kwon \\
  Department of Bio and Brain Engineering\\
  KAIST\\
  \texttt{cyclomon@kaist.ac.kr} \\
  \And
  Jong Chul Ye \\
  Kim Jaechul Graduate School of AI\\
  KAIST\\
  \texttt{jong.ye@kaist.ac.kr} \\
}

\begin{document}

\maketitle
\begin{center}
\centering
\includegraphics[width=1.0\linewidth]{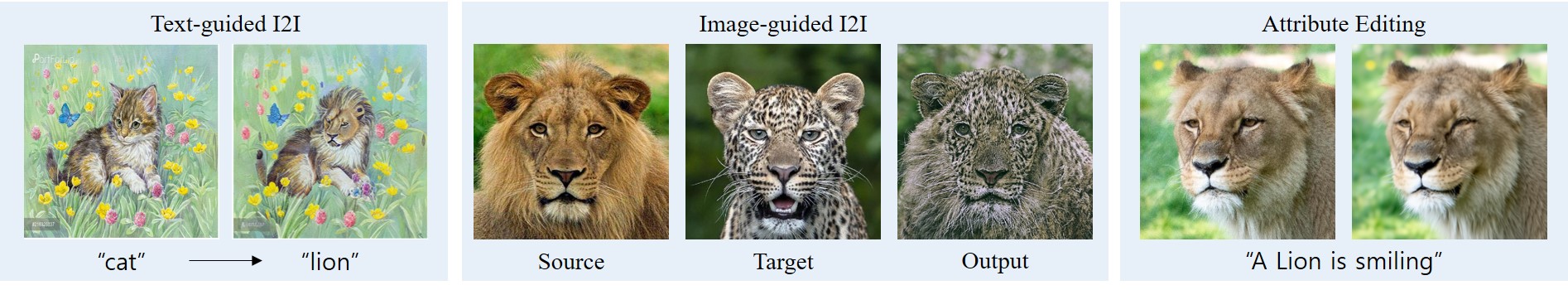}
\captionof{figure}{Image translation examples by our proposed method. Our model can  generate high-quality translation and editing outputs using both text and image conditions. }
\label{fig:first}
\end{center}

\begin{abstract}
Diffusion models have shown significant progress in image translation tasks recently. However, due to their stochastic nature, there's often a trade-off between style transformation and content preservation. Current strategies aim to disentangle style and content, preserving the source image's structure while successfully transitioning from a source to a target domain under text or one-shot image conditions. Yet, these methods often require computationally intense fine-tuning of diffusion models or additional neural networks.
To address these challenges, here we present an approach that guides the reverse process of diffusion sampling by applying
{\em asymmetric} gradient guidance. This results in quicker and more stable image manipulation for both text-guided and image-guided image translation. Our model's adaptability allows it to be implemented with both image- and latent-diffusion models. Experiments show that our method outperforms various state-of-the-art models in image translation tasks.
\end{abstract}

\section{Introduction}

Image translation is a task that aims to convert an image from a source domain to a target domain while preserving the structural attributes of the source image. In the past, image translation methods based on Generative Adversarial Networks(GAN) \cite{gan} were mainly used, ranging from unimodal image-to-image translation (I2I) \cite{distanceGAN,gcgan,cut,HnegSRC} to multimodal I2I models ~\cite{munit,starganv2}. While these methods showed good performance, they had the disadvantage of being unable to perform image-to-image translation for arbitrary domains and conditions. 

Recently, various methods have been proposed to solve this problem.
Particularly notable are the methods that manipulate the generation process of a pre-trained generative model by applying modifications to generate images with desired conditions. This approach has the advantage of using the superior generation performance of the pre-trained model while being able to generate the desired images. Specifically, previous methods mainly used pre-trained GAN models to enable image manipulation~\cite{interface,ganspace}. Recently, the use of large foundation models, especially text-to-image embedding models such as CLIP~\cite{clip_rad}, has been introduced for image editing based on text conditions~\cite{styleclip,nada}. However, these methods show limited performance due to the limitations of GAN models, such as limited domains and the cumbersome process of embedding real images into the latent space.

With the development of diffusion models and the combination of text encoder models, text-to-image diffusion models~\cite{imagen,ldm} have been developed, and research on image translation and image editing using diffusion models has become more active  \cite{cgd,glide}. In particular, various methods have been proposed based on cross-attention with text conditions~\cite{p2p,pnp}, allowing for image-to-image translation. However, the text-to-image model requires an inversion process that accurately embeds the source image to latent space, which consumes additional time for optimization.
Moreover, they have difficulty applying to image translation cases where the source image is provided as condition.

To address this,
a recent approach of DiffuseIT~\cite{diffuseit} uses gradient guidance during the generation process to preserve the structural attributes of the source image while disentangling and changing the target style.
While these models have shown good performance, they still requires extensive use of slicing ViT network \cite{splice}
and its computationally expensive gradient computation.


To address the limitations of existing methods, here we  present a novel sampling approach using so-called
Asymmetric Gradient Guidance (AGG), which amalgamates the strengths of various models. 
Specifically, the gradient update by AGG is composed of two step:
a single MCG framework  \cite{mcg} to compute the initial update, which is followed by
computationally efficient Decomposed Diffusion Sampling (DDS) update from the denoised signal by Tweedie' fomula, which can
be efficiently   computed by Adam optimizer.
Furthermore, in contrast to DiffuseIT that relies on slicing ViT based structural preservation loss,
 we use much simpler structural  regularization term  using the intermediate product stored during the forward DDIM step.
 Combined with simpler regularization loss and efficient gradient update, our
 method can provide much faster and effective image translation under text- and one-shot image conditions.
Moreover, this integration  provides a significant advantage as it enables our sampling approach to operate in both traditional image diffusion models and latent diffusion models \cite{ldm}.

Our contribution can be summarized as follows:
\begin{itemize}
  \item We propose Asymmetric Gradient Guidance method, which 
  is composed of two step:
a single MCG framework  \cite{mcg} followed by
computationally efficient Adam optimization from the denoised signal by Tweedie' fomula.
  \item We introduce a simpler structural loss function which utilizes the intermediate products saved during the DDIM forward step, allowing for faster and efficient gradient
  guidance in both image diffusion models and latent diffusion models.

 \item Our method is so efficient and  flexible that  it can be integrated with arbitrary loss functions, making it suitable for a wide range of applications such as text-guided image translation, image-guided appearance transfer, artistic style transfer, image editing, etc.

\end{itemize}

\section{Related Works}

Recently, the combination of pre-trained GAN models and the CLIP model has been proposed for various applications 
such as naive generation~\cite{vqclip}, style transfer with fine-tuning the models \cite{nada}, and text-guided editing \cite{styleclip}. However, generating images through GAN has limitations such as restricted domains and poor performance.
 With the development of diffusion-based models, recent attempts combined the CLIP and the diffusion model for various tasks such as generating text-conditioned images ~\cite{cgd,glide} or editing images using mask regions ~\cite{blended}. Some methods involve tuning the entire or part of the model to transform images~\cite{diffusionclip,asyrp}. Methods that can transform images based on text conditions without training but only through the sampling process \cite{diffuseit,zst} have also  demonstrated impressive performance.


Independent from text-conditioned image manipulation, several researchs have been conducted on image manipulation with target image conditions. Initially, image style transfer models using feature tensor modulation \cite{sanet,wct2,strotss} were developed. Subsequently methods adapting GAN models \cite{oneshotclip,fsga,mind} showed good performance. Recent models can transfer the appearance of one-shot images based on ViT feature \cite{splice}, and subsequent studies inspired by this have successfully combined diffusion models to transfer appearance \cite{diffuseit}. 


\section{Background}


\noindent\textbf{Denoising Diffusion Probabilistic Models (DDPM)}
DDPM\cite{ddpm} is trained to estimate the added noise in the noisy image $\x_t$  produced by Markovian forward process :
\begin{equation}\label{eq:eq1}
    q(\x_{T}|\x_0):=\prod_{t=1}^{T}q(\x_t|\x_{t-1}),\quad \mbox{where}\quad
    q(\x_t|\x_{t-1}):=\Nc(\x_t;\sqrt{1-\beta_t}\x_{t-1},\beta_t\Ib),
\end{equation}
Here, the noise schedule $\beta_t$ is an increasing sequence of $t$, with $\bar\alpha_t := \prod_{i=1}^t \alpha_t,\, \alpha_t := 1 - \beta_t$. 
  For the reverse process, with the learned noise prediction network $\boldsymbol{\epsilon}_\theta$, we can predict the noisy sample of previous timestep $\x_{t-1}$:   
\begin{align}\label{eq:reverse}
    \x_{t-1}=\frac{1}{\sqrt{\alpha_t}}\Big{(}\x_t-\frac{1-\alpha_t}{\sqrt{1-\bar{\alpha}_t}}\boldsymbol{\epsilon}_\theta(\x_t,t)\Big{)}+\sigma_t\boldsymbol{z},
\end{align}
where $\z\sim\Nc(0,\Ib)$ and $\sigma^2$ is variance which is set as $\beta$ in DDPM.
With iterative process, we can sample the image $\x_0$ from initial sample $\x_{T}\sim\Nc(0,\Ib)$.

\noindent\textbf{Denoising Diffusion Implicit Models (DDIM)}
Since the diffusion models requires large number of steps, the full sampling process is slow. To address the problem, DDIM~\cite{ddim} proposes another method by re-defining the DDPM formulation as non-Markovian process. Specifically, the reverse step process is described as :
\begin{align}\label{eq:ddim}
    \x_{t-1}=\underbrace{ \sqrt{\bar\alpha_{t-1}}\hat \x_{0,t}(\epsilonb_{\theta}(\x_t,t))}_\text{Denoise} + 
    \underbrace{D_t(\epsilonb_{\theta}(\x_t,t))}_\text{Noise} + \sigma_t \z_t,  \quad 
 \end{align}
 where
 \begin{align}
\label{eq:tweedie}
  \hat \x_{0,t}(\epsilonb_{\theta}(\x_t,t)) := \frac{\x_t-\sqrt{1-\bar\alpha_t}\epsilonb_\theta(\x_t,t)}{\sqrt{\bar\alpha_t}},&\quad
  D_t(\epsilonb_{\theta}(\x_t,t)):=\sqrt{1-\bar\alpha_{t-1}-\sigma_t^2}\epsilon_{\theta}(\x_t,t)
\end{align}   
If $\sigma_t=0$, the reverse process is fully deterministic, leading to a faster sampling.

\section{Main Contribution}


The recent focus  of the conditional diffusion researches is how to incorporate
the conditioning gradient during the reverse sampling using \eqref{eq:reverse} or \eqref{eq:ddim}.
This is because for a given loss function $\ell(\x)$, a direct injection of the gradient of the
loss computed at $\x_t$ produces inaccurate gradient guidance.
To address this, various methods have been developed. In this following, we first provide a
brief review of them and present a novel Asymmetric Gradient guidance (AGG) by taking the best of 
the existing methods.

\subsection{Gradient guidance: a review}

\begin{figure}[t!]
    \centering
    \includegraphics[width=1.0\linewidth]{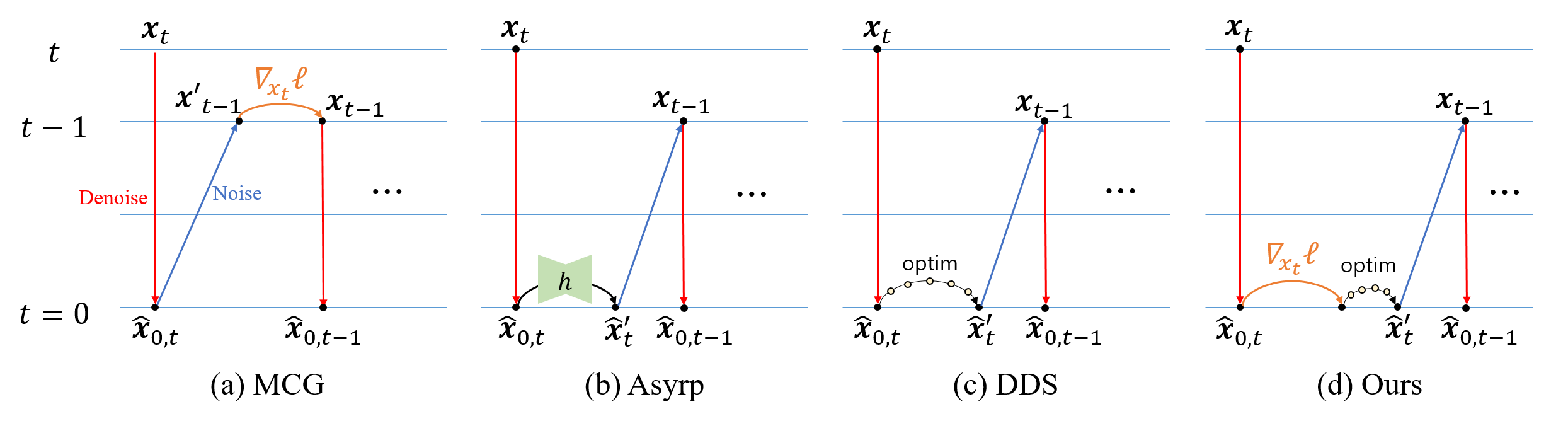}
    \caption{Various gradient guidance scheme. (a) MCG utilizes manifold gradient that ensures the update sample to stay on the correct noisy manifold.
  (b) Asyrp updates the diffusion model  for denoising formula with h-space manipulation. (c) DDS use optimization based update to guide the denoised image. (d) Our proposed method is composed of one-step MCG and Tweedie's formula followed by  additional DDS optimization. 
}
    \label{fig:fig_method}
    \end{figure}

\noindent\textbf{MCG.}
 The manifold constraint gradient (MCG)~\cite{mcg} begins by projecting the noisy image $\x_t$ onto the estimated clean image $\hat \x_{0,t}$
 using the Tweedie's formula in \eqref{eq:tweedie}. 
 Specifically, the sampling involves guiding the noisy image $\x_{t-1}$ as follows:
\begin{align}\label{eq:mcg}
\x'_{t-1} \leftarrow \x_{t-1}-\nabla_{\x_t}\ell(\hat \x_{0,t}(\x_t)),
\end{align}
where $\x_t$ is the initially obtained next-step sample as per \eqref{eq:ddim}, and $\nabla{\x_t}\ell(\hat \x_{0,t}(\x_t))$ is the gradient of the loss $\ell$, calculated with the denoised image $\hat \x_{0,t}(\x_t)$ with respect to $\x_t$. 
It was shown in \cite{mcg} that the MGC gradient enforces the update to stay on the original
noisy manifold (see Fig.~\ref{fig:fig_method} (a)).

\noindent\textbf{Asyrp.} Asymmetric Reverse Sampling (Asyrp)  \cite{asyrp}
converts the reverse sampling step in \eqref{eq:ddim} to the following:
\begin{align}\label{asyrp}
    \x_{t-1} = \sqrt{\bar\alpha_{t-1}}\hat \x_{0,t}(\tilde\epsilonb_{\theta}(\x_t,t)) + D_t(\epsilonb_{\theta}(\x_t,t)) + \sigma_t \z_t,
\end{align}
where
 \begin{align} 
 \Tilde{\epsilonb}_{\theta}(\x_t,t) = {\epsilonb}_{\theta}(\x_t,t) + \Delta\epsilonb_t
 \end{align}
To compute the update of the diffusion model $\Delta\epsilonb_t$, they introduced a concept of $h$-space, which is to manipulate the bottleneck feature of score network $\epsilonb_\theta$. Therefore, they set   $\hat \x_{0,t}(\Tilde{\epsilonb}_{\theta})$ as $\hat \x_{0,t}({\epsilonb}_{\theta}(\x_t|\Delta h_t,t))$ in \eqref{asyrp}, which is an estimated clean image from score network with the manipulated $h$-space (see Fig.~\ref{fig:fig_method}(b)).

\noindent\textbf{DDS.} 
In Decomposed Diffusion Sampling (DDS)~\cite{dds}, the following optimization
problem is solved after projecting the sample on the clean manifold through Tweedie's formula:
\begin{align}\label{eq:delta}
\Delta\x_0 =    \arg\min_\Delta \ell(\hat \x_{0,t}+\Delta), 
\end{align}
which results in the following modified reverse sampling step:
\begin{align}\label{dds}
    \x_{t-1} = \sqrt{\bar\alpha_{t-1}}\left(\hat \x_{0,t}(\epsilonb_{\theta}(\x_t,t)) +\Delta\x_0\right)+ D_t(\epsilonb_{\theta}(\x_t,t)) + \sigma_t \z_t,
\end{align}
It was shown \cite{dds} that under some regularity condition the conjugate gradient (CG)-based update $\Delta \x_0$ still ensures
the updated sample $\hat \x_{0,t}+\Delta\x_0$ to stay on the clean manifold, so that subsequent noising process transfers the sample to a correct noisy manifold
 (see Fig.~\ref{fig:fig_method} (c)).

\subsection{Asymmetric Gradient Guidance}

\noindent\textbf{Key observation.}
Although aforementioned approaches appear seemingly different, there are fundamental connection between them.
Specifically, by translating $\Delta\x$ of \eqref{eq:delta} back to the noisy data space of $\x_t$, the authors in \cite{universal} observed that the original forward model 
$\x_t = \sqrt{\bar\alpha_{t}}\x_0 +\sqrt{1-\bar\alpha_t}\epsilonb_t$ can be converted to
\begin{align}
\x_t = \sqrt{\bar\alpha_{t}}(\hat\x_{0,t}+\Delta \x_0) +\sqrt{1-\bar\alpha_t}\tilde\epsilonb_t,\quad 
\end{align}
where the perturbed diffusion model is given by
\begin{align}
 \tilde\epsilonb_t := \epsilonb_t - \sqrt{\bar\alpha_t/(1-\bar\alpha_t)}\Delta \x_0
\end{align}
Therefore, the diffusion model update by Asyrp and the sample
domain update by DDS are equivalent up to a scaling factor.
Furthermore, using the method of the variation, $\Delta\x_0$ in \eqref{eq:delta} can be
represented by $\Delta\x_0 =-\eta \nabla_{\x_t} \ell(\hat \x_{0,t}(\x_t))$, leading to 
\begin{align}
 \tilde\epsilonb_t := \epsilonb_t +s_t \nabla_{\x_t} \ell(\hat \x_{0,t}(\x_t))
\end{align}
for some $s_t$, suggesting the condition based guidance of the diffusion model.
The previous work \cite{universal} sets $s_t=\sqrt{1-\bar\alpha_t}$.

%

\noindent\textbf{Proposed method.}
Given the equivalence between the sample update and diffusion model update, we can
derive a general update formula for the conditional diffusion:
\begin{align}\label{general}
    \x_{t-1} = \sqrt{\bar\alpha_{t-1}}\hat{\x}'_t + D_t(\epsilonb_{\theta}(\x_t,t)) + \sigma_t \z_t, 
 \end{align}
 where the following two forms are interchangely used for $\hat \x_t'$:
 \begin{align}\label{eq:xdot}
    \hat{\x}'_t = \begin{cases}
    \hat \x_{0,t}({\epsilonb}_t) - \nabla_{\x_t}\ell(\hat \x_{0,t} (\x_t)) \\
    \hat \x_{0,t}(\tilde{\epsilonb}_t), \quad \mbox{where}\quad \tilde{\epsilonb}_t: = {\epsilonb}_t- s_t\nabla_{\x_t}\ell(\hat \x_{0,t}(\epsilonb_t))
    \end{cases}
\end{align}
Therefore, the remaining question is the computation of the MCG gradient $\nabla_{\x_t}\ell(\hat \x_{0,t} (\x_t))$, which 
requires a backpropagation.
The key idea of the DDS is that for general linear inverse problems,
when the underlying solution space lies in the Kyrov subspace,
a few step of standard CG update starting from a denoised sample
still ensures the update samples to live in the clean manifold.
Therefore, the computationally expensive backpropagation for MGC can be bypassed \cite{dds}.

Unfortunately, we found that DDS approaches does not provide a stable solution when the loss function is originated from
text-driven and image-driven image translation of our interest, as the associated inverse problem
cannot be represented as a linear inverse problem.
However, applying the MCG at each iteration is computationally expensive.
To address this problem, we propose the following hybrid update of $\hat{\x}_t'$ for \eqref{general}:
\begin{align*}
\hat{\x}_t' = \bar\x_t' + \arg\min_\Delta \ell(\bar\x_t'+\Delta),& \quad \mbox{where} \quad \bar\x_t' := \hat \x_{0,t}(\tilde{\epsilonb}_t)
\end{align*}
where the loss minimization in the first term is done without MGC gradient using  Adam optimizer,
and the perturbed model output $\tilde\epsilonb_t$ is computed by \eqref{eq:xdot}.
This implies that for image translation tasks one step MGC gradient  is necessary for the successive
DDS type optimization to ensures the updated sample to live on the correct manifold. We call this as {\em Asymmetric Gradient Guidance (AGG)}.

AGG can be applied not only to image diffusion models but also to latent diffusion models (LDM) \cite{ldm}. In this case, instead of using $\x_t$ in the image diffusion, we utilize an embedded latent $\z_t$ after the encoder. 
\begin{figure}[t]
    \centering
    \includegraphics[width=0.8\linewidth]{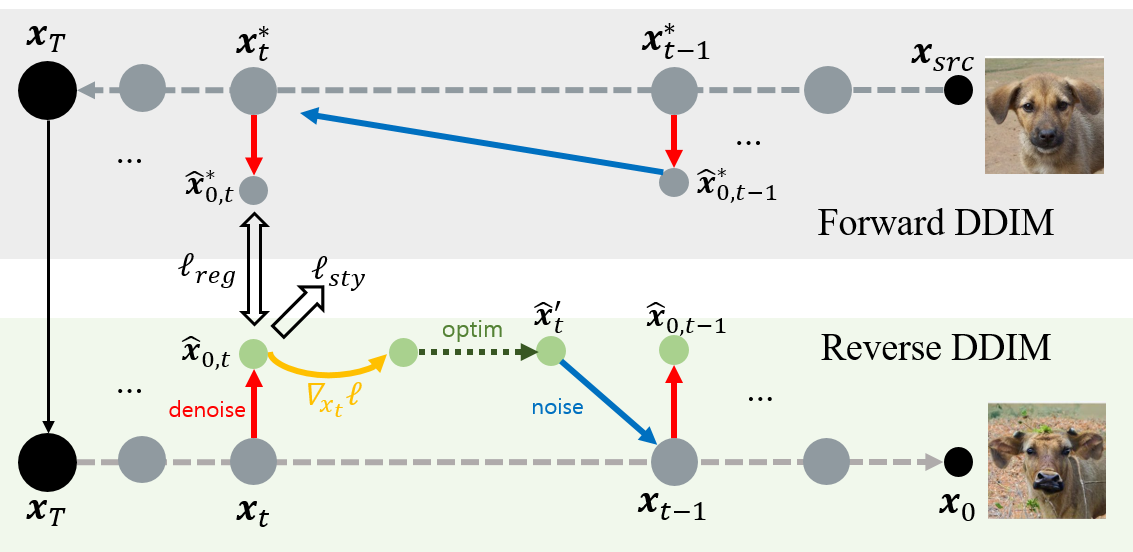}
    \caption{Detailed sampling strategy of our proposed method. We start our sampling process with encoding the real image $\x_{src}$ with forward DDIM to $\x_T$. During forward step, we save the intermediate estimated clean images $\hat{\x}^*_{0,t} = \hat{\x}^*_{0,t}(\epsilonb_t)$ to use them as target for regularization. With initial $\x_{T}$, we guide the reverse sampling through loss function calculated with $\hat \x_{0,t} =  \hat \x_{0,t}(\epsilonb_t)$. We manipulate $\hat{\x}_t$ with the calculated gradient and additional optimization step.}
    \label{fig:fig_method2}
\end{figure}

\subsection{Loss function}
For image translation, our loss function is composed of two parts: style loss $\ell_{sty}$ for image domain change, and the regularization loss $\ell_{reg}$  for preserving the source image structure. 
\begin{align}
    \ell_{total}= \lambda_{sty} \ell_{sty} + \lambda_{reg}\ell_{reg},
\end{align}
where we can control the content preservation strength with controlling parameter $\lambda_{reg}$. 

As for $\ell_{sty}$, we leveraged the style loss functions used in DiffuseIT \cite{diffuseit}. 
Specifically, for text-guided I2I, we use embedding similarity loss using pre-trained CLIP model, whereas for image-guided I2I, we use [CLS] token matching loss using DINO ViT\cite{dino}.

For structural preservation, DiffuseIT utilize a powerful content preservation loss using ViT. However, we propose a simpler yet more efficient regularization approach.  
%
%
Specifically, we propose  a  regularization scheme of  the reverse step using the pre-calculated intermediate samples from DDIM forward process. 
As shown in Fig.~\ref{fig:fig_method2}, in the forward sampling of DDIM, we can deterministically calculate the intermediate samples by modifying the reverse sampling in  \eqref{eq:ddim}. With changing the timestep $t$ and setting $\sigma_t=0$ such as: 
\begin{align}\label{eq:ddim_forward}
    {\x}^*_{t+1}={ \sqrt{\bar\alpha_{t+1}}\hat{\x}^*_{0,t}(\epsilonb_t)} + {\sqrt{1-\bar\alpha_{t+1}}\epsilonb_t}, 
\end{align}
where ${\x}^*_{t}$ is calculated noisy image from source image ${\x}_{src}$, and $\hat{\x}^*_{0,t}(\epsilonb_t)$ is estimated denoised image calculated using Tweedie's formula in \eqref{eq:tweedie}. Furthermore, rather than saving for all $t$,  we save $\hat{\x}^*_{0,t}(\epsilonb_t)$ for $T-t_{edit}<t$, in which $t_{edit}$ is point where we stop the gradient guidance.
Using the saved intermediate denoised samples, our regularization loss is defined as: 
$$\ell_{reg}=d(\hat{\x}^*_{0,t}({\epsilon}^{\theta*}_t),\hat \x_{0,t}({\epsilon}^{\theta}_t))$$ in where we use $l_1$ loss for the distance metric $d(\cdot,\cdot)$. 

Our regularization has two main effects. Firstly, it provides a general effect of guiding the output to follow the attributes of the source image by preventing significant deviations. Secondly, since the guidance target is also a denoised image at a specific timestep $t$, it helps to prevent the manipulated clean image $\hat{\x}'_t$ from deviating from original manifold. This constraint leads to an improvement in the quality of the generation output, as demonstrated in the ablation study included in our appendix.

\section{Experimental Results}
\paragraph{Experimental Details}
For implementation, we refer to the official source code of DiffuseIT~\cite{diffuseit}. For experiment, we use unconditional score model pre-trained with Imagenet 256$\times$256 resolution datasets~\cite{diffusionbeat}. In all the experiments, we used diffusion step of $T = 60$.  We only used guided reverse step until $t>T-t_{edit}$. We set $t_{edit}=20$ for image diffusion model. The generation process takes 15 seconds per image in single RTX 3090 including forward and reverse DDIM in text-guided I2I. Our detailed experimental settings are elaborated in Appendix.

\begin{figure}[t!]
\centering
    \includegraphics[width=0.9\linewidth]{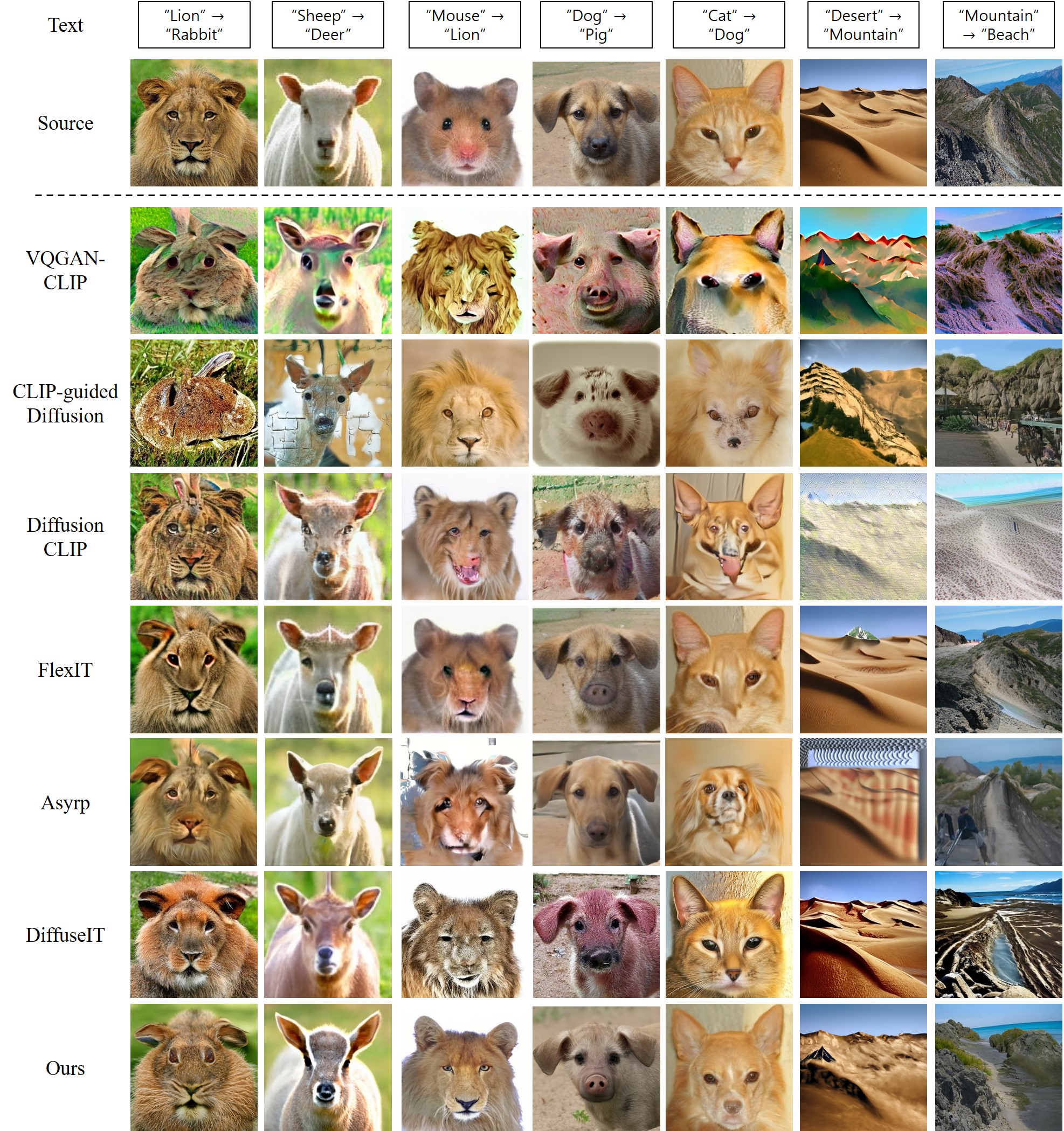}
    \caption{Qualitative comparison of text-guided  translation on \textit{Animals} and \textit{Landscape} dataset. Our model shows better perceptual quality than the baselines.  }
    \label{fig:fig_ani}
    \end{figure}

\begin{figure}[!t]

\begin{center}
\resizebox{0.8\textwidth}{!}{
\begin{tabular}{@{\extracolsep{5pt}}cccccccc@{}}
\hline
\multirow{2}{*}{\textbf{Method}}  & \multicolumn{3}{c}{\textbf{Animals}} &\multicolumn{3}{c}{\textbf{Landscapes}} & \textbf{Time}\\

\cline{2-4} 
\cline{5-7}
\cline{8-8}
 & SFID$\downarrow$& CSFID$\downarrow$&LPIPS$\downarrow$& SFID$\downarrow$& CSFID$\downarrow$&LPIPS$\downarrow$ & sec \\

\hline
VQGAN-CLIP &30.01&	65.51 &	 0.462 &	33.31& 82.92  & 0.571 & 8s\\
CLIP-GD &12.50&53.05 &0.468 &18.13& 62.19& 0.458 & 25s\\
DiffusionCLIP& 25.09& 66.50& 0.379 &29.85&76.29 & 0.568& 70s\\
FlexIT &32.71&	57.87 &	0.215 &	18.04 &60.04 & 0.243& 26s\\
Asyrp &31.41&	89.60 &	0.338 &	 23.65 & 74.32 &  0.457& 65s\\
DiffuseIT &9.98&	41.07	& 0.372	&16.86& 54.48 & 0.417& 40s\\
Ours &{9.64}&	{38.12}	& {0.336}	&{14.33}& {54.11} & {0.414}& 15s \\
\hline
\end{tabular}
}
\end{center}
\captionof{table}{ Quantitative comparison for the text-guided image translation. Our model outperforms baselines in overall scores for both of \textit{Animals} and \textit{Landscapes} datasets with faster sampling.}
\label{table:text0}
\end{figure}

\paragraph{Text-guided Image Translation}
In order to assess the effectiveness of our text-guided image translation approach, we performed comparisons against the state-of-the-art baseline models which use image diffusion models. The chosen baseline methods were recently introduced models that utilize pre-trained CLIP for text-guided image manipulation : VQGAN-CLIP \cite{vqclip}, CLIP-guided diffusion (CGD) \cite{cgd}, DiffusionCLIP \cite{diffusionclip}, FlexIT \cite{flexit}, Asyrp \cite{asyrp}, and DiffuseIT \cite{diffuseit}. To ensure consistency, we referred to the official source codes of all the baseline methods.


Following the same experimental protocol used in DiffuseIT, we show quantitative and qualitative evaluation on natural image translation. For testing our translation performance, we use two different datasets: animal faces \cite{animal} and landscapes \cite{landscape}. The animal face dataset contains 14 classes of animal face images, and the landscapes dataset consists of 7 classes of various natural landscape images. 

For quantitative metrics, we used three different metrics following the baseline DiffuseIT \cite{diffuseit} : simplified FID (SFID) \cite{sfid}, class-wise SFID (CSFID), and LPIPS. For overall image quality measurement, we used SFID score. For caterogy-wise quality check, we show CSFID. To measure the input-output structural consistency, we calculated LPIPS distance between source and outputs.
Further experimental settings can be found in our Supplementary Materials.


In Table \ref{table:text0}, we show the quantitative comparison results. 
Our model demonstrated superior performance in image quality measurement using SFID and CSFID compared to all baseline methods. Regarding content preservation measured by the LPIPS score, our approach achieved the second-best performance. The FlexIT model obtained the highest score in LPIPS since it was directly trained with LPIPS loss. However, an excessively low LPIPS value is undesirable as it indicates a failure in achieving proper semantic changes. 
With respect to the translation time, training-based methods Asyrp and DiffusionCLIP require longer time. Comparing with other sampling-based methods, our model showed relatively faster sampling speed. Considering the output quality, our model is the most efficient method for I2I.

In qualitative comparison in Fig.~\ref{fig:fig_ani}, our results accurately capture the semantic features of the target texts while preserving the content. On the other hand, baseline models suffered from poor sample quality or failure to change the domain.
To further evaluate the perceptual quality of generated samples, we show user study results in our Appendix. 

\begin{figure}[t!]
\centering
    \includegraphics[width=0.9\linewidth]{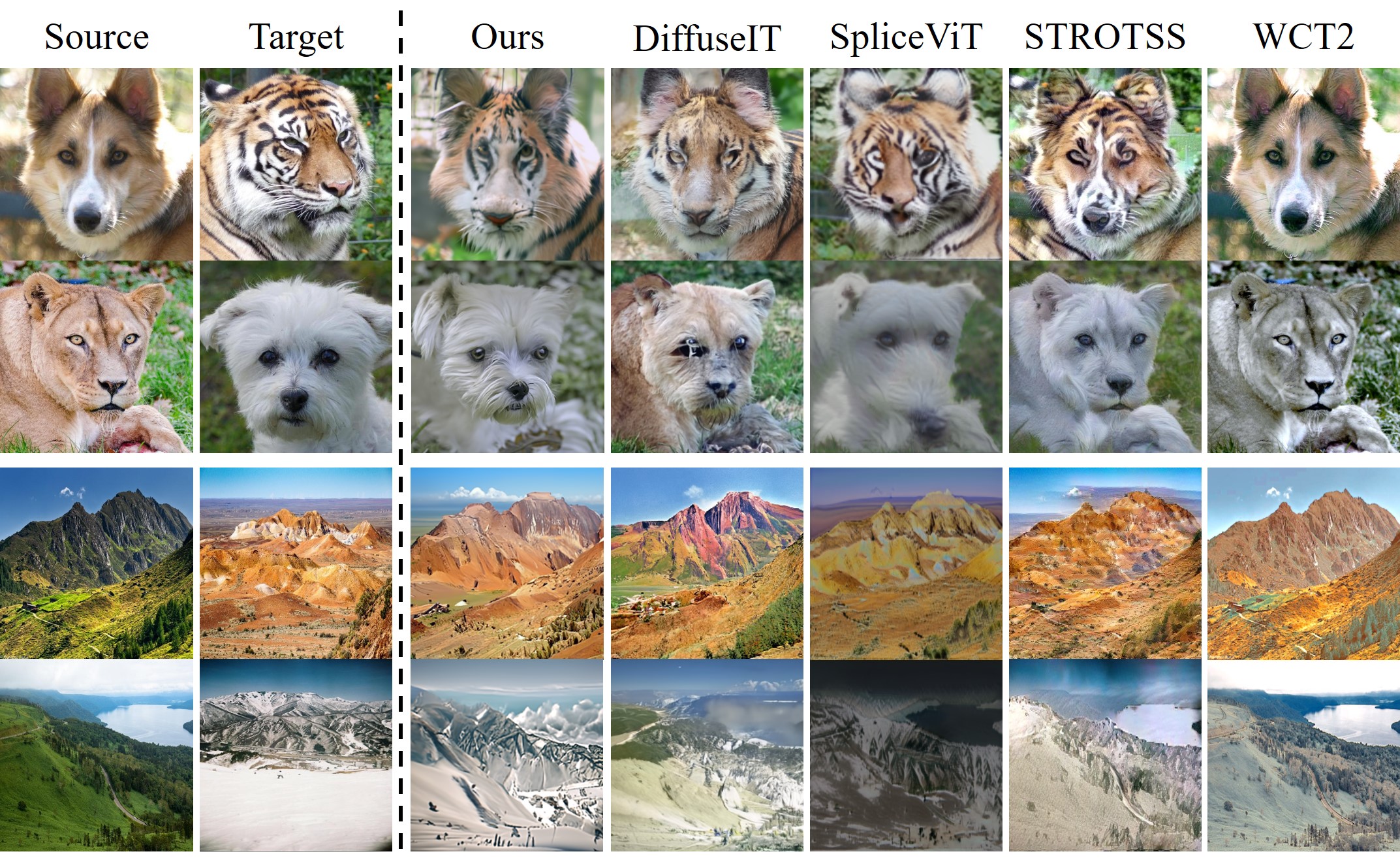}
    \caption{Qualitative comparison of image-guided image translation. Our results have better perceptual quality than the baseline outputs. }
    \label{fig:fig_img}
    \end{figure}

\paragraph{Image-guided Image Translation}
To assess the image translation guided by target images, we conducted comparison experiments to evaluate its performance. Specifically, we compared our model against appearance transfer models such as DiffuseIT \cite{diffuseit}, Splicing ViT \cite{splice}, STROTSS \cite{strotss}, and style transfer methods WCT2 \cite{wct2}.

Figure \ref{fig:fig_img} presents the qualitative comparison results of the image-guided translation task. Our model successfully generated outputs that accurately captured the semantic styles of the target images while preserving the content from the source images. Conversely, the other models exhibited severe content deformations or inadequately reflected the semantic styles. Additionally, we show a user study to evaluate the overall perceptual quality in Supplementary Materials.

\begin{figure}[t!]
\centering
    \includegraphics[width=0.9\linewidth]{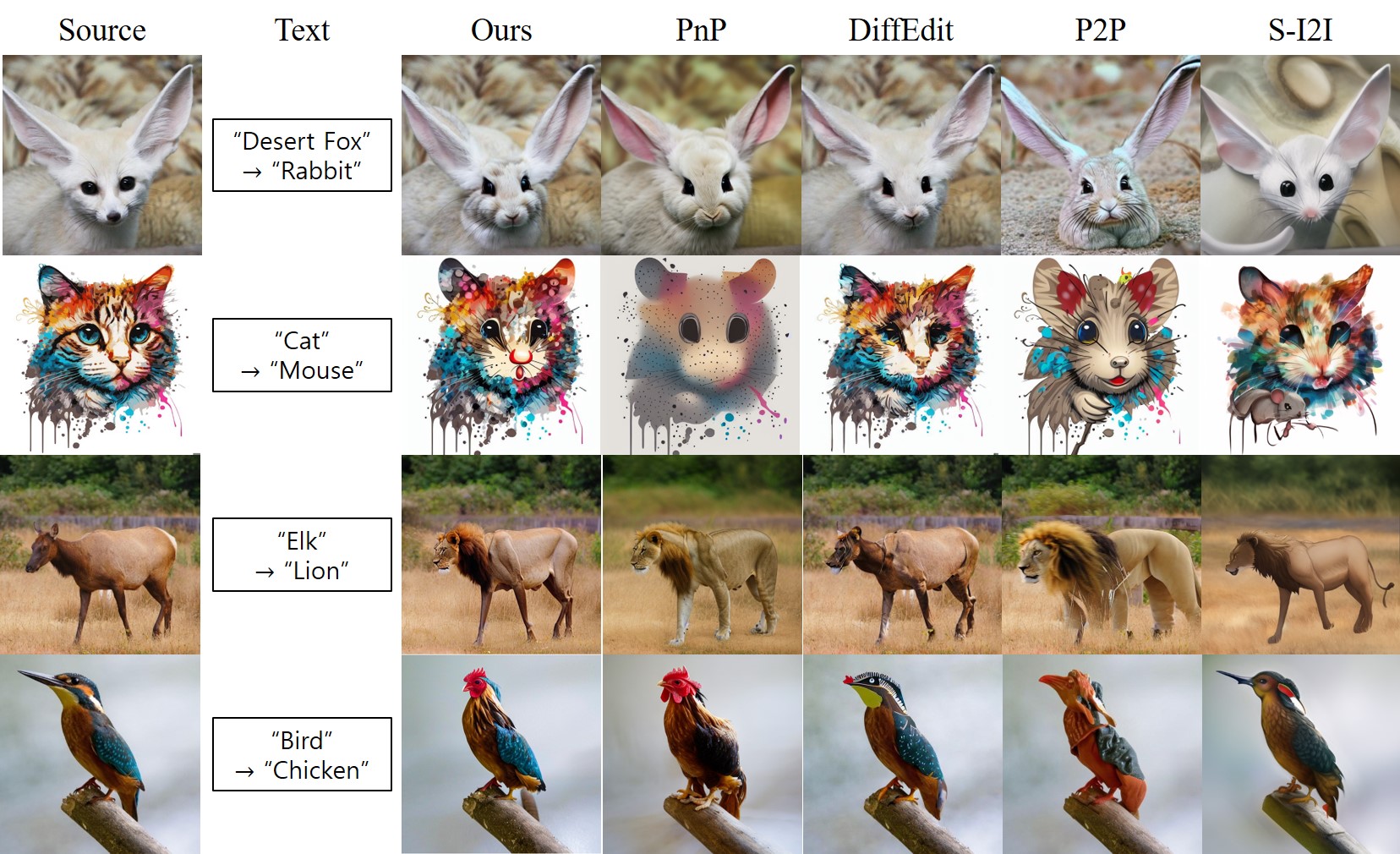}
    \caption{Qualitative comparison of text-guided  translation with latent diffusion model.  Our model generates realistic samples that reflects the both of source image attribute and target text condition, with better perceptual quality than the baselines. }
    \label{fig:fig_ldm}
    \end{figure}



\paragraph{Experiment on Latent Diffusion Models}
As baseline models, we used stable diffusion I2I (S-I2I), prompt-to-prompt (P2P) \cite{p2p}, plug-and-play diffusion model (PnP) \cite{pnp}, and Diffedit \cite{diffedit}, which are real image manipulation methods. Except for the simple forward-reverse process of S-I2I, all other methods required real image inversion, which took more than 2 minutes. In contrast, our method only utilized the sampling process, which took less than 20 seconds. Detailed time comparisons and experimental details on latent diffusion model can be found in Appendix.

Fig.~\ref{fig:fig_ldm} presents the results obtained using our latent diffusion model. Our model successfully achieved semantic changes in images within a short time. In contrast, the baselines exhibited limitations such as the inability to achieve proper semantic changes or failure to preserve the shape of the source image. Recent models, particularly PnP diffusion, showed pleasing quality in natural image translation but failed when applied to artificial images.

\begin{figure}[t!]
\centering
    \includegraphics[width=0.90\linewidth]{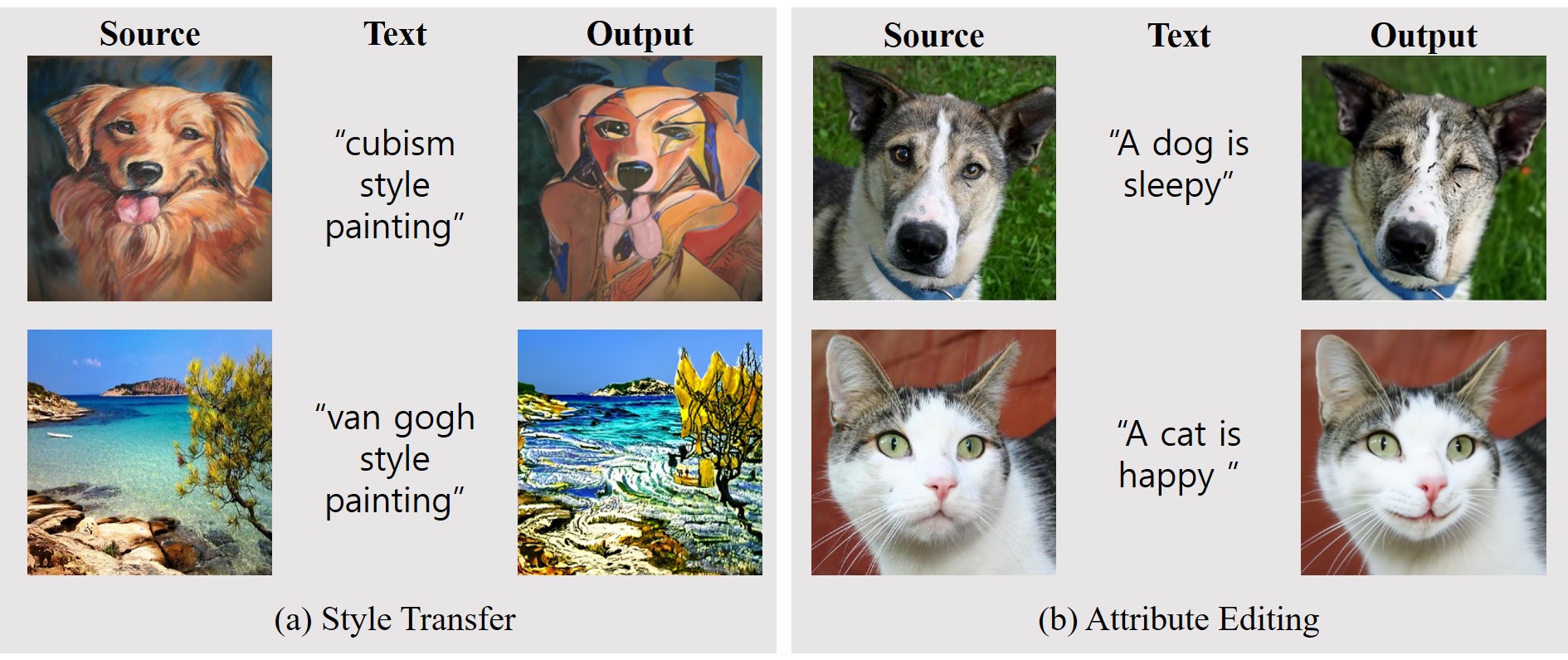}
    \caption{Various applications of our proposed method. Our method can be used for overall style transfer and attribute editing.  }
    \label{fig:fig_other}
    \end{figure}

\paragraph{Other applications}
By adjusting the regularization loss $\lambda_{reg}$, we can expect various effects. Reducing the loss strength allows for overall style changes, resembling style transfer, while increasing the loss strength and making all steps deterministic enables editing of the source image attributes. In Fig.~\ref{fig:fig_other}, we show the results of style transfer and image editing.  
More samples and experimental details are provided in our Supplementary Material.

\section{Societal Impacts, Limitations, and Reproducibility}

\begin{wrapfigure}{r}{0.3\linewidth}
\vspace{-0.4cm}
    \centering
\includegraphics[width=1.0\linewidth]{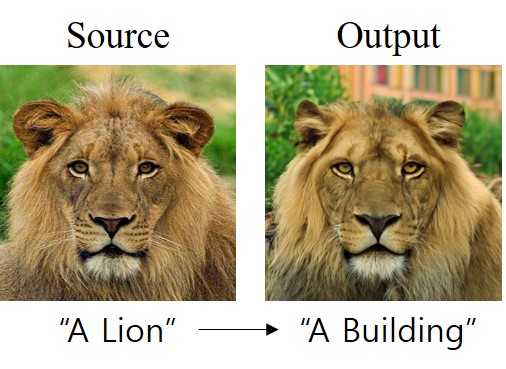}
\caption{Example of failure case.}
\label{fig:failure}
\vspace{-0.4cm}
\end{wrapfigure}

\textbf{Societal impacts.} Our method can contribute to society by empowering fields such as entertainment and the arts through conditional generation. However, there are also negative aspects, such as the malicious use of the method to create deepfakes. These concerns can be addressed through regulatory measures.

\textbf{Limitations.} Despite the good performance in image translation, it has limitations. Since we use pre-trained CLIP model, there is a problem when a text condition has large distance from source image in CLIP space (e.g., lion -> building) as shown in Fig.~\ref{fig:failure}. This limitation can be addressed in the future with the development of better text embedding models.

\textbf{Reproducibility.} In supplementary materials, we provide an anonymous link containing codes, algorithms and hyper-parameters.

\section{Conclusion}
In this paper, we propose a novel method that utilizes asymmetric gradient guidance during the reverse sampling process to transform the source image to the target domain. Our proposed method enables more stable and faster image translation compared to existing reverse sampling approaches. Our experimental results demonstrate improved performance compared to recently proposed state-of-the-art image translation models, achieving successful image-to-image (I2I) translation in both image- and latent-diffusion models. Furthermore, our model is not confined to semantic I2I, with showing variety of applications such as style transfer and image editing.

\small
\bibliographystyle{unsrt}

\bibliography{neurips}

\newpage
\appendix

\section{Experiment on Latent Diffusion Models}
As described in our main paper, out method can be easily adapted to the recent Latent Diffusion Model (LDM)~\cite{ldm}.
With encoded latent $\z$, we can obtain noisy latent $\z_t$ with the same process of the forward DDIM process of image diffusion models. 
With re-iterating our sampling process, we can define our update formula for latent diffusion model:
\begin{align}\label{general_ldm}
    \z_{t-1} = \sqrt{\bar\alpha_{t-1}}\hat{\z}'_t + D_t(\epsilonb_{\theta}(\z_t,t,c_{trg})) + \sigma_t \epsilon, 
 \end{align}
 where the following two forms are interchangely used for $\hat \z_t'$:
 \begin{align}\label{eq:xdot_ldm}
    \hat{\z}'_t = \begin{cases}
    \hat \z_{0,t}({\epsilonb}_t) - \nabla_{\z_t}\ell(\hat \z_{0,t} (\z_t)) \\
    \hat \z_{0,t}(\tilde{\epsilonb}_t), \quad \mbox{where}\quad \tilde{\epsilonb}_t: = {\epsilonb}_t- s_t\nabla_{\z_t}\ell(\hat \z_{0,t}(\epsilonb_t))
    \end{cases}
\end{align}
where $c_{trg}$ is the text embedding of target text $t_{trg}$, and $\epsilon\sim\Nc(0,\Ib)$, $\epsilonb_t = \epsilonb(z_t,t,c_{trg})$, and $\nabla_{\z_t}\ell(\hat \z_{0,t} (\epsilonb_t))$ refers the MCG gradient.

Similar to the image diffusion case, we apply additional DDS process for hybrid update of $\hat{\z}'_t$ for Eq. \eqref{general_ldm} with the following formula: 
\begin{align*}
\hat{\z}_t' = \bar\z_t' + \arg\min_\Delta \ell(\bar\z_t'+\Delta),& \quad \mbox{where} \quad \bar\z_t' := \hat \z_{0,t}(\tilde{\epsilonb}_t).
\end{align*}
The argmin operation in the right hand side is done without MGC gradient using  Adam optimizer,
and the perturbed model output $\tilde\epsilonb_t$ is computed by \eqref{eq:xdot_ldm}.

In latent diffusion case, we can optionally leverage the cross attention map of text conditions. Especially, we obtain the semantic mask for the specific word in sentence. With following the recent model~\cite{p2p}, we can obtain the saliency map by aggregating the cross attenton map $A$ of specific word. With applying threshold to saliency map, we can obtain semantic mask $M$, which cover the  targe object we want to translate. 
More specifcally, with obtained mask, we can use differentiated guided sampling strategy, in which we apply guided sampling until different timesteps with respect to masks: 
 \begin{align}
    \hat{\z}'_t = \begin{cases}
    \hat{\z}'_t , \quad \mbox{where}\quad  T > t > T-t_{edit1}\\
     (1-M)*\hat{\z}'_t  + M*\hat{\z}_{0,t}({\epsilonb}_t), \quad \mbox{where}\quad T-t_{edit1} \geq t > T-t_{edit2} \\
     \hat{\z}_{0,t}({\epsilonb}_t), \quad \mbox{where}\quad T-t_{edit2} \geq t
    \end{cases}
\end{align}
With the manipulated denoised image, we can predict next sample output with  the formula \eqref{general_ldm}.

For the loss function, we only use regularization loss term in our latent diffusion case, as the model already incorporates text condition. 
Specifically, in the forward sampling of DDIM, we can deterministically calculate the intermediate samples such as: 
\begin{align}\label{eq:ddim_forward_ldm}
    {\z}^*_{t+1}={ \sqrt{\bar\alpha_{t+1}}\hat{\z}^*_{0,t}(\epsilonb(z_t,t,c_{src}))} + {\sqrt{1-\bar\alpha_{t+1}}\epsilonb(z_t,t,c_{src})}, 
\end{align}
where ${\z}^*_{t}$ is calculated noisy image from source latent obtained from encoder ${\z}_{src} = Enc(\x_{src})$, and $\hat{\z}^*_{0,t}(\epsilonb(z_t,t,c_{src}))$ is estimated denoised image calculated using Tweedie's formula.
Using the saved intermediate denoised samples, our regularization loss is defined as: 
$$\ell_{reg}=d(\hat{\z}^*_{0,t}({\epsilonb}(z^*_t,t,c_{src})),\hat \z_{0,t}({\epsilonb}(z_t,t,c_{trg})))$$ in where we use $l_1$ loss for the distance metric $d(\cdot,\cdot)$. 

\section{Experimental Details}
For experiment hyperparameters, we set $\lambda_{reg}=200$ and $\lambda_{sty}=200$ in MCG gradient calculation for image diffusion models. For additional DDS optimization part, we used the same loss function as MCG, using Adam optimizer with learning rate of 0.02. In image diffusion model case, we used only one or two steps of DDS optimization, as it showed marginal improvement with more steps. For all of the experiments, we give moderate random stochastic noise with setting $\sigma_t=0.8$, as it showed better perceptual quality.

For style loss, we used the loss function proposed in DiffuseIT~\cite{diffuseit}. More specifically, for text-guided image translation, our style loss is defined as:
\begin{align}
\ell_{CLIP}(\x; \d_{trg}, \x_{src},\d_{src}) := - \mathrm{sim}(\v_{trg},\v_{src})
\end{align}
where
\begin{align}
\v_{trg}:= E_T(\d_{trg}) + \lambda_i E_I(\x_{src}) - \lambda_s E_T(\d_{src}),&\quad \v_{src}:=  E_I(\mathrm{aug}(\x)) 
\end{align}
where $E_T$ and $E_I$ is text and image encoder, respectively, and $\d_{src}$,$\d_{trg}$,$\x_{src}$ represents source text, target text and source image, respectively.  In this case, $\x$ is estimated denoised image;
$\mathrm{aug(\cdot)}$ denotes the augmentation for preventing adversarial artifacts from CLIP. For parameters, we use exactly same values proposed in DiffuseIT.

For image-guided image translation, we also use proposed loss in DiffuseIT, which is defined as:
\begin{align}
\ell_{sty}(\x_{trg},\x) = ||\e_{[CLS]}^L(\x_{trg})-\e_{[CLS]}^L({\x})||_2 + \lambda_{mse}||\x_{trg}-{\x}||_2.
\end{align}
where $\e_{[CLS]}^L$ denotes the last layer  [CLS] token of DINO ViT model.

For experiments of latent diffusion models, we used public stable diffusion model with version v1.5. We used total timesteps $T=50$ for all experiments. For hyperparameters, we set $\lambda_{reg}=0.1$ in MCG gradient calculation. For additional DDS optimization, we used Adam with weight decay (AdamW) optimizer with learning rate of 0.01. We used 10 DDS optimization steps per each timestep. For editing timesteps, we set $t_{edit2} = 40$, and  $t_{edit1}$ is varied between $[10:20]$. We provide the source code in our official repository\footnote{\url{https://github.com/submissionanon18/AGG}}.

\section{User Study}

\begin{figure}[!t]
\begin{minipage}{.48\linewidth}
\begin{center}
\resizebox{1\textwidth}{!}{
\begin{tabular}{@{\extracolsep{5pt}}cccc@{}}
\hline

\cline{1-4} 
 \textbf{Method}& Text$\uparrow$& Realism$\uparrow$&Content$\uparrow$ \\

\hline
VQGAN-CLIP &2.20&	1.61&	 1.77 \\
CLIP-GD &2.47& 1.47& 1.38 \\
DiffusionCLIP& 1.73& 1.93& 2.43 \\
FlexIT &2.12&	3.36 &	3.69 \\
Asyrp &2.79&	2.69&	2.77 \\
DiffuseIT &2.81&3.71	& 3.54	\\
Ours &{4.07}&	{4.14}	& {4.23} \\
\hline
\end{tabular}
}
\end{center}
\captionof{table}{ User study for the text-guided image translation on image diffusion models. Our model outperforms baselines in overall scores.}
\label{table:user_text}
\end{minipage}
\hfill
\begin{minipage}{.48\linewidth}
\begin{center}
\resizebox{1\textwidth}{!}{
\begin{tabular}{@{\extracolsep{5pt}}cccc@{}}
\hline

\cline{1-4} 
 \textbf{Method}& Style$\uparrow$& Realism$\uparrow$&Content$\uparrow$ \\

\hline

WCT2& 2.28& 4.46& 4.28 \\
STROTSS & 3.28&	3.58 &	3.72 \\
SpliceViT & 2.85&	1.92&	2.39\\
DiffuseIT &3.07& 2.07	& 2.70	\\
Ours &{4.21}&	{3.79}	& {3.85}  \\
\hline
\end{tabular}
}
\end{center}
\captionof{table}{ User study for the image-guided image translation on image diffusion models. Our model outperforms baselines in overall scores.}
\label{table:user_image}
\end{minipage}
\end{figure}

\begin{figure}[!t]
\begin{minipage}{.48\linewidth}
\begin{center}
\resizebox{1\textwidth}{!}{
\begin{tabular}{@{\extracolsep{5pt}}cccc@{}}
\hline

\cline{1-4} 
 \textbf{Method}& Text$\uparrow$& Realism$\uparrow$&Content$\uparrow$\\

\hline

S-I2I& 2.84&  3.47& 3.30 \\
P2P & 4.19&	3.82 &	3.59 \\
DiffEdit & 2.74&	2.61&	3.35 \\
PnP &3.94& 2.83&  2.43	\\
Ours &{3.82}&	{3.89}	& {4.05} \\
\hline
\end{tabular}
}
\end{center}
\captionof{table}{ User study for the text-guided image translation on latent diffusion models.}
\label{table:user_ldm}
\end{minipage}
\hfill
\begin{minipage}{.46\linewidth}
\resizebox{0.48\textwidth}{!}{
\begin{tabular}{@{\extracolsep{5pt}}cc@{}}
\hline

\cline{1-2} 
 \textbf{Method}& Time\\

\hline

WCT2&  0.18s\\
STROTSS &  53s\\
SpliceViT &  25m 30s\\
DiffuseIT & 37s\\
Ours & 15s \\
\hline
\end{tabular}
}
\hfill
\resizebox{0.43\textwidth}{!}{
\begin{tabular}{@{\extracolsep{5pt}}cc@{}}
\hline

\cline{1-2} 
 \textbf{Method}& Time\\

\hline

S-I2I&  5s\\
P2P &  2m 8s\\
DiffEdit &  12s\\
PnP & 2m 36s\\
Ours & 17s \\
\hline
\end{tabular}
}
\captionof{table}{Time comparison results for evaluating the efficiency of our model. }\label{table:time}
\end{minipage}

\end{figure}

For evaluation of perceptual quality of our proposed model, we conducted user study with following similar protocol to baseline model~\cite{diffuseit}.

First, we conducted a user study to evaluate the performance of our proposed and baseline models in text-guided image translation. The study included 49 images generated from 7 different text conditions. Participants were asked questions regarding three aspects: 1) whether the outputs accurately conveyed the semantic meaning of the target text (Text-match), 2) the realism of the generated images (Realism), and 3) the preservation of the content information from the source images (Content). A total of 20 users from different age groups (20s and 50s) were randomly recruited for the study. The participants were provided with the questions using Google Form and were asked to rate the answers on a scale from 1 to 5, where 1 represented "Very Unlikely" and 5 represented "Very Likely."
In the user study on image-guided translation, we generated 20 distinct images. We then followed a similar procedure as the text-guided image translation study, with the exception of the question content. Participants were asked to provide feedback in three categories: 1) whether the outputs accurately conveyed the semantic meaning of the target style image (Style-match), 2) the realism of the generated images (Realism), and 3) the preservation of content information from the source images (Content).
For the user study involving the Latent diffusion model, we presented 20 different images and gathered user opinions. In this case, we employed the exact same set of questions used in our text-guided translation study on image diffusion models.

Based on the criterion,  we first evaluated the performance of text-guided image translation applied to image diffusion model. In Table~\ref{table:user_text}, we show the user study results. In overall scores including text-matching, realism, and content preservation, our model scored the best among other baseline models. 
Also we show the user study result of image-guided image translation in Table~\ref{table:user_image}. Our model scored the best in style-matching score and the second best in realism and content preservation score. In WCT2, the model shows the best score in realism and content preservation, but severely low score in style matching score. This indicates that the model could not properly change the source images into the target domain.
Finally, we show the user study results on latent-diffusion based models. In Text-matching semantic score, baseline P2P, PnP and our model all showed decent scores. However, ours showed the best score in realism and content-preservation score. As already shown in our main script, baseline of PnP and P2P shows good performance in generating images which has proper semantic of target texts, but they show degraded performance in preserving the attribute of original images. 

\section{Computation Time Comparison}
To compare the computation times taken for image translation, we show the comparison results in Table~\ref{table:time}. In image-guided translation, our model shows the second-best performance in translation time. For baseline models of Stable diffusion I2I and DiffEdit, the consumed times are relatively low as they are only applied to sampling steps. However, they suffer from degraded performance for image translation. In advanced method of P2P and PnP, they show comparable performance in image translation, but they require longer time (>2min) since they require additional process of inversion. For P2P, we incorporated the standard inversion process of null-text inversion~\cite{mokady2022nulltext}. For PnP, we used the DDIM-inversion process with following the official source code.  

\section{Ablation Study}

\begin{figure}[t!]
\centering
    \includegraphics[width=1.0\linewidth]{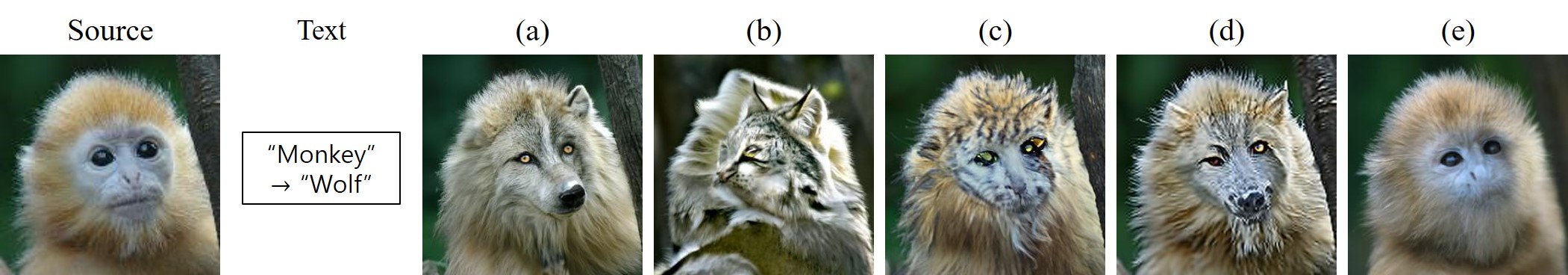}
    \caption{Qualitative ablation study results. (a) Our best setting. (b) Outputs without proposed $\ell_{reg}$. (c) Generated images only using DDS. (d) Outputs with Vit-based regularization (e) Outputs without our proposed asymmetric update rule. }
    \label{fig:fig_ablation}
    \end{figure}

\begin{table}[!t]

\begin{center}

\begin{tabular}{@{\extracolsep{5pt}}cccc@{}}
\hline
\multirow{2}{*}{\textbf{Settings}}  & \multicolumn{3}{c}{\textbf{Animals}}\\

\cline{2-4} 
 & SFID$\downarrow$& CSFID$\downarrow$&LPIPS$\downarrow$ \\
\hline
Symmetric update (e) &4.04 & 60.87 & 0.260 \\
ViT Loss instead of $\ell_{reg}$ (d)& 21.24& 50.04& 0.312  \\
Only DDS (c)& 29.66&	71.01&	0.460  \\
No $\ell_{reg}$ (b) &14.65 &	41.17	& 0.448 \\
Ours (a) &9.64&	38.12	& 0.336	\\
\hline
\end{tabular}

\end{center}
\caption{Quantitative comparison of ablation studies. }
\label{table:abl}
\end{table}

To verify the effectiveness of our proposed components, we show the ablation study results. In Fig. \ref{fig:fig_ablation}, we show the results using different experimental settings. (a) Firstly, we show the generated outputs with our best setting. (b) To evaluate out proposed regularization loss, we generated results without using our proposed $\ell_{reg}$. (c) We generated results using only DDS optimization. (d) We show the generated results using ViT regularization loss proposed in DiffuseIT~\cite{diffuseit}, instead of our proposed $\ell_{reg}$. (e) We show the generated images using symmetric update rule in which we apply MCG gradient to both of denoised and noise components. 

In case of (b), we can see the content structure is severely degraded. In (c) and (d), we can observe that the results maintain the structure of source images, but they suffer from unwanted artifacts. In (e), the model could not change the semantic of source image to target texts. These pattern are also exhibited in our quantitative ablation study in Table~\ref{table:abl}. Without asymmetric update in (e), SFID and LPIPS scores are extremely low, while class-wise FID shows relatively high score. This indicates that the model hardly changed the semantic of source images.  With using ViT loss instead of our proposed regularization loss, we can see the LPIPS score is lower than ours, but the image quality is degraded due to artifacts. For (c) and (b), the models failed in image translation with degraded scores compared to our model. In overall, our proposed setting shows the best quality.

\section{Algorithms}
For detailed explanation, we show algorithms in Algorithm \ref{alg:alg1}.
\begin{algorithm}[t]
    \caption{Image translation: given a diffusion score model $\epsilon_{\theta}(\x_t,t)$, CLIP model, and VIT model}
     \textbf{Input:} source image $\x_{src}$, diffusion steps $T$, editing timestep $t_{edit}$ target text $\d_{trg}$, source text $\d_{src}$ or target image $\x_{trg}$\\
     \textbf{Output:} translated image $\hat{\x}$ which has semantic of $\d_{trg}$ (or $\x_{trg}$) and content of $\x_{src}$ \\
     $\x_{src} = \x^*_0$
    \begin{algorithmic}[1]

        \ForAll{$t$ from 0 to $T$}
            \State $\epsilon^* \leftarrow \boldsymbol{\epsilon}_{\theta}(\x^*_t,t)$
            \State $\hat{\x}^*_{0,t} \leftarrow \frac{\x^*_t-\sqrt{1-\bar\alpha_t}\epsilon^*}{\sqrt{\bar\alpha_t}}$
            \State save $\hat{\x}^*_{0,t}$
            \State $\x^*_{t+1}= \sqrt{\bar\alpha_{t+1}}\hat{\x}^*_{0,t} + 
    \sqrt{1-\bar\alpha_{t+1}-\sigma_t^2}\epsilon^* + \sigma_t \z_t $
        \EndFor 
        \State $\x_T \leftarrow \x^*_T$
        \ForAll{$t$ from $T$ to 0}
            
            \State $\epsilon \leftarrow \boldsymbol{\epsilon}_{\theta}(\x_t,t)$
            \State $\hat{\x}_{0,t} \leftarrow \frac{\x_t-\sqrt{1-\bar\alpha_t}\epsilon}{\sqrt{\bar\alpha_t}}$
            \If{$T$> $t$ >$T-t_{edit}$}
                \If{text-guided}
                    \State $\ell_{total}(\hat{\x}_{0,t}) \leftarrow [\ell_{CLIP}(\hat{\x}_{0,t};\d_{trg},\x_{src},\d_{src}) + d(\hat{\x}_{0,t},\hat{\x}^*_{0,t}) ]$
                \ElsIf{image-guided}
                    \State $\ell_{total}(\hat{\x}_{0,t}) \leftarrow [\ell_{sty}(\hat{\x}_{0,t};\x_{trg}) + d(\hat{\x}_{0,t},\hat{\x}^*_{0,t})]$
                \EndIf
    
                \State $\Tilde{\epsilon} \leftarrow \epsilon - s_t\nabla_{\x_t}\ell_{total}(\hat{\x}_{0,t})$
                \State $\bar{\x}_{0,t} \leftarrow \hat{\x}_{0,t}(\Tilde{\epsilon})$
    
                \If{use-DDS} 
                    \State $\hat{\x}'_{0,t} \leftarrow \bar{\x}_{0,t} + \text{arg}\min_\Delta\ell_{total}(\bar{\x}_{0,t} + \Delta)$
                \Else
                    \State $\hat{\x}'_{0,t} \leftarrow \bar{\x}_{0,t}$
                \EndIf
            
                \State $\z \sim \mathcal{N}(0,\mathbf{I})$
                \State $\x_{t-1}= \sqrt{\bar\alpha_{t-1}}\hat{\x}'_{0,t} + 
        \sqrt{1-\bar\alpha_{t-1}-\sigma_t^2}\epsilon + \sigma_t \z_t $
            \Else
                \State $\z \sim \mathcal{N}(0,\mathbf{I})$
                \State $\x_{t-1}= \sqrt{\bar\alpha_{t-1}}\hat{\x}_{0,t} + 
            \sqrt{1-\bar\alpha_{t-1}-\sigma_t^2}\epsilon + \sigma_t \z_t $
            \EndIf
        \EndFor 
        \State \textbf{return} $\x_{-1}$
    \end{algorithmic}
    \label{alg:alg1}
    \end{algorithm}

\section{Additional Experiments}

\begin{figure}[t!]
\centering
    \includegraphics[width=1.0\linewidth]{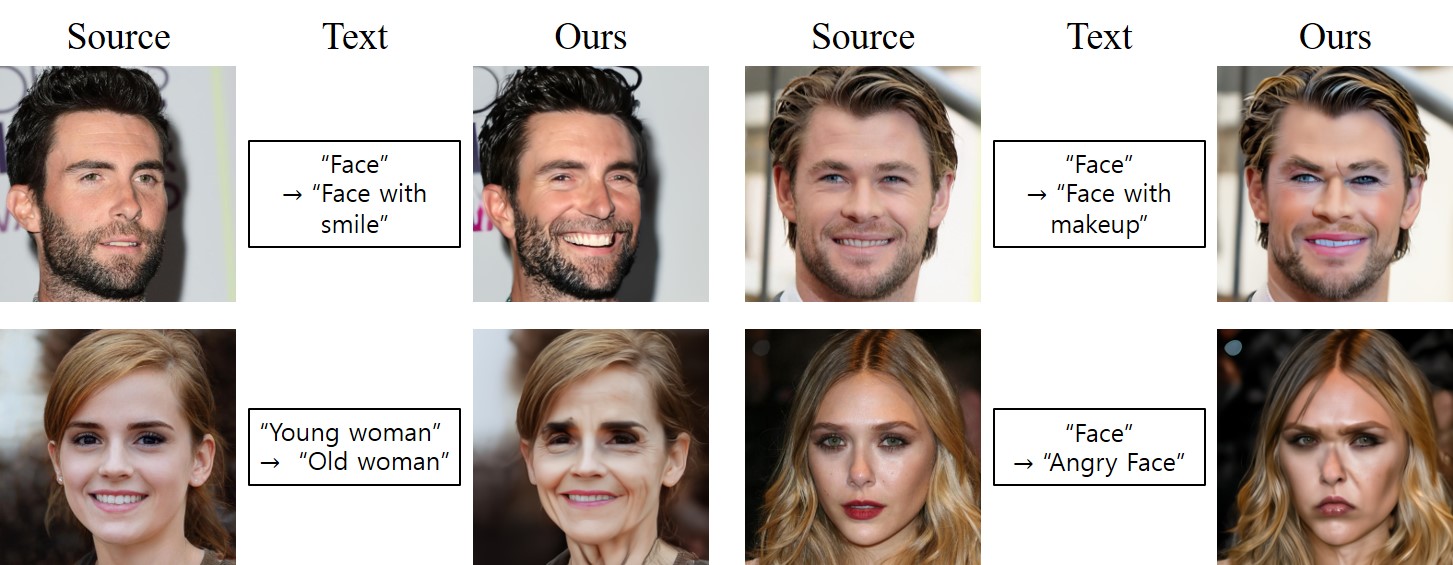}
    \caption{Qualitative results on human face image translation. }
    \label{fig:fig_face}
    \end{figure}

In this section, we show additional generated samples. First, we show the image-translation results using image diffusion models pre-trained on human face dataset. In Fig.~\ref{fig:fig_face}, the results show that our model can be easily applied to human face editing. We can change the face attribute using target text conditions, while preserting the original facial features. 

For the additional applications, we can replace our regularization loss into other losses such as cross-entropy of semantic segmentation model or object detection model. In this case, we use large number of DDS steps more than 2,000. During DDS step, we used resampling strategy in which we apply one reverse and forward DDIM step after each 100 DDS steps with following the strategy of ~\cite{universal}. We applied our guided sampling to initial 3 to 5 timesteps, as the initial stage of diffusion sampling plays major role in the overall generated contents. 
For semantic segmantation, we combined pre-trained model from previous work~\cite{seg_pre}. In object detection model, we also used pre-trained model ~\cite{detection}. 

\begin{figure}[t!]
\centering
    \includegraphics[width=1.0\linewidth]{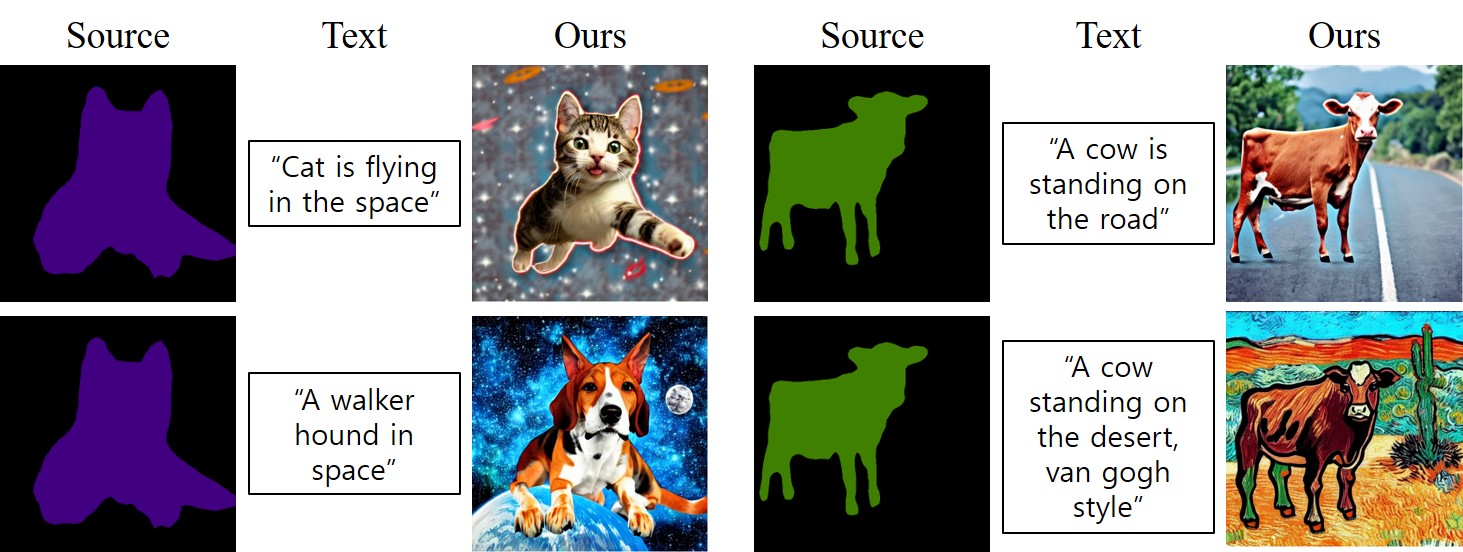}
    \caption{Qualitative results of generated images using pre-trained semantic segmentation models. }
    \label{fig:fig_seg}
    \end{figure}

\begin{figure}[t!]
\centering
    \includegraphics[width=1.0\linewidth]{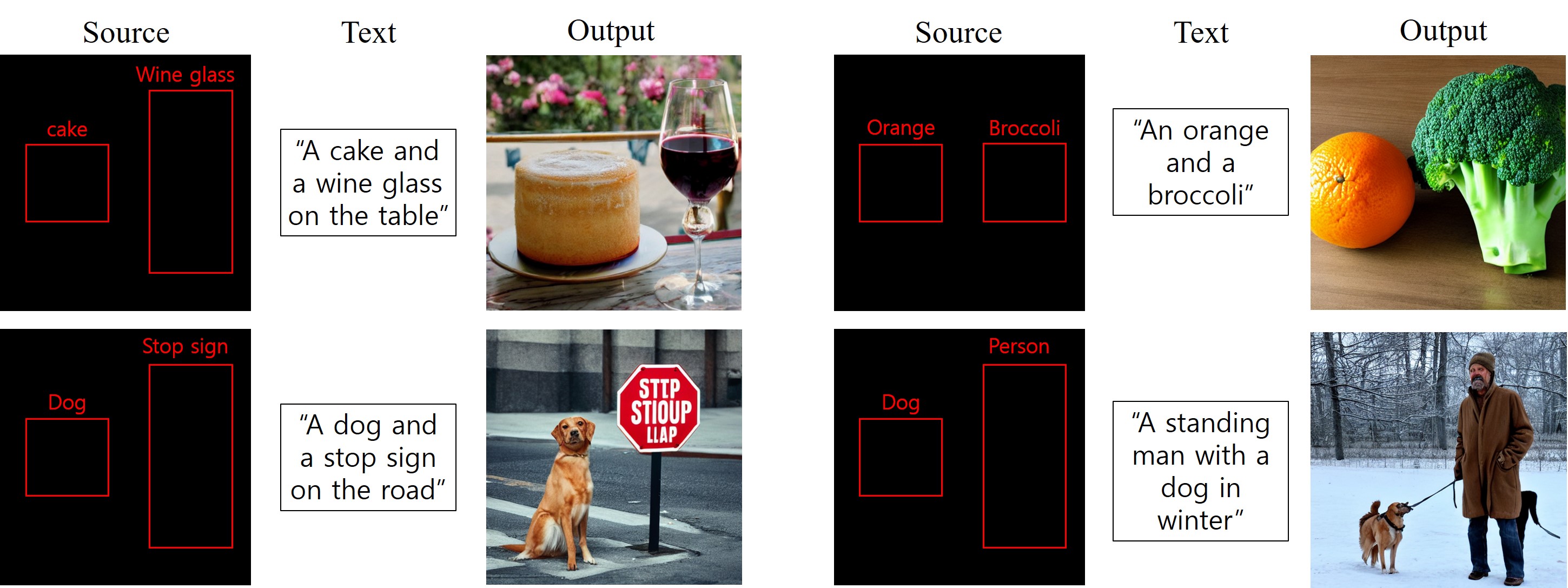}
    \caption{Qualitative results of generated images using pre-trained object detection model. }
    \label{fig:fig_obj}
    \end{figure}

In Fig.~\ref{fig:fig_seg}, we show the generated output using segmentation model. We can generate text-conditioned images in which the image components are aligned with semantic segmentation maps. In Fig. \ref{fig:fig_obj}, we also show the outputs using pre-trained object detection model. We can generate images in which the objects positions are aligned with given bounding box information.

\begin{figure}[t!]
\centering
    \includegraphics[width=1.0\linewidth]{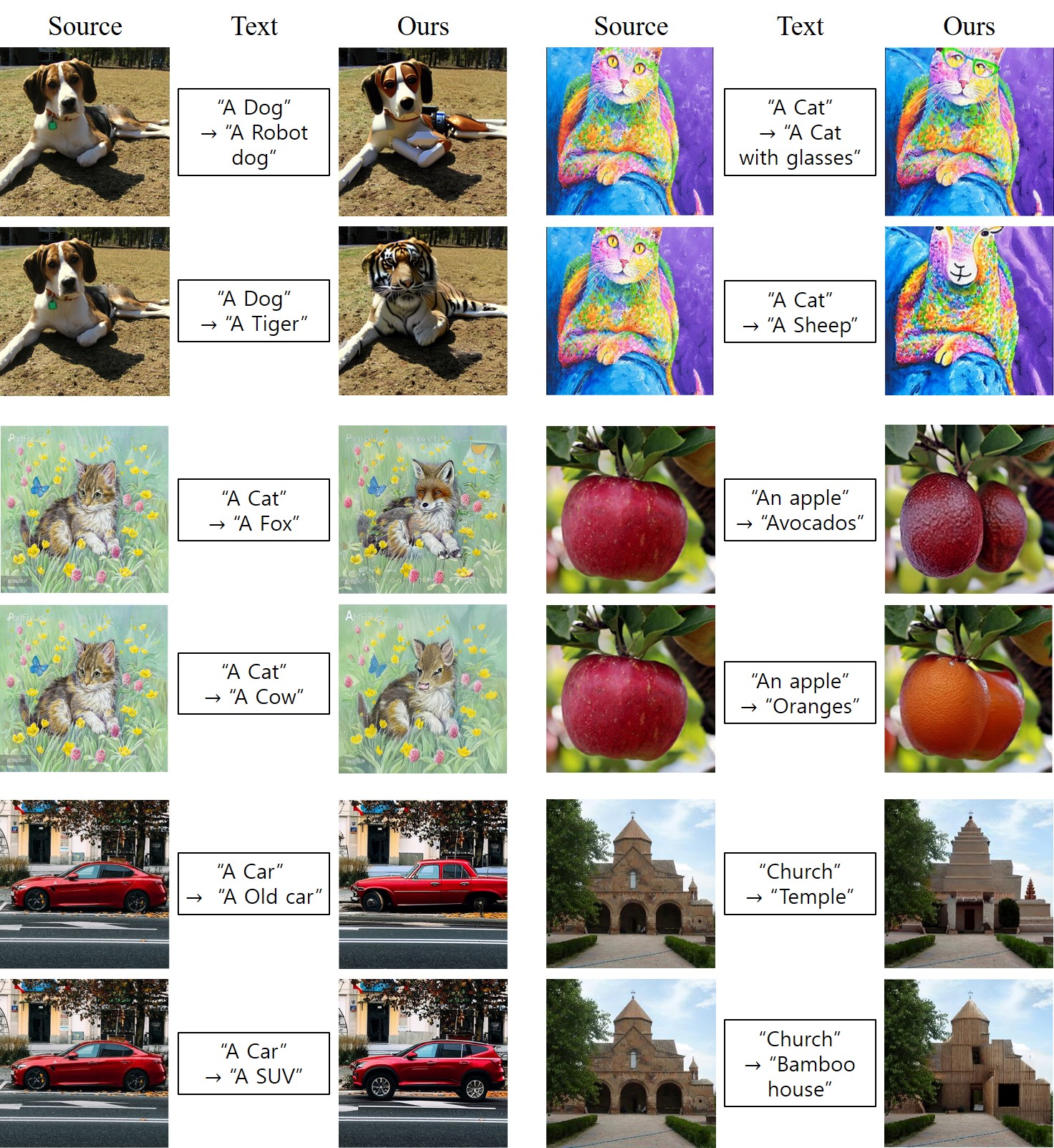}
    \caption{Qualitative results of generated images using latent diffusion model. }
    \label{fig:fig_ldm_add}
    \end{figure}

To show more image translation examples using our method, we show additional samples generated from latent diffusion models in Fig.~\ref{fig:fig_ldm_add}.
\newpage
\newpage



\end{document}